\documentclass[conference,a4paper]{IEEEtran}

\IEEEoverridecommandlockouts                              
\usepackage{graphicx}                                                          

\begin{document}
\title{\LARGE \bf
Crack-pot: Autonomous Road Crack and Pothole Detection}
\author{\IEEEauthorblockN{Sukhad Anand\IEEEauthorrefmark{1},
Saksham Gupta\IEEEauthorrefmark{2}, Vaibhav Darbari\IEEEauthorrefmark{3} and
Shivam Kohli\IEEEauthorrefmark{4}}
\IEEEauthorblockA{
Email: \IEEEauthorrefmark{1}sukhad.anand@gmail.com,
\IEEEauthorrefmark{2}saksham@vt.edu,
\IEEEauthorrefmark{3}vaibhavdarbari@gmail.com,
\IEEEauthorrefmark{4}kohlishivam5522@gmail.com}}

\maketitle
\thispagestyle{empty}
\pagestyle{empty}

\begin{abstract}
With the advent of self-driving cars and autonomous robots, it is imperative to detect road impairments like cracks and potholes and to perform necessary evading maneuvers to ensure fluid journey for on-board passengers or equipment. We propose a fully autonomous robust real-time road crack and pothole detection algorithm which can be deployed on any GPU based conventional processing boards with an associated camera. The approach is based on a deep neural net architecture which detects cracks and potholes using texture and spatial features. We also propose pre-processing methods which ensure real-time performance. The novelty of the approach lies in using texture-based features to differentiate between crack surfaces and sound roads. The approach performs well in large viewpoint changes, background noise, shadows, and occlusion. The efficacy of the system is shown on standard road crack datasets. 
\end{abstract}

\section{INTRODUCTION}\label{intro}
Asphalt surfaced roads and pavements are constantly subjected to heavy traffic and changing weather conditions leading to their gradual deterioration and degradation resulting in traffic delays, compromising commuter safety and reducing efficiency.The deformations often surface in the form of potholes and cracks which pose a threat to the vehicles in absence of timely evasive maneuvers. This task of dodging road malformations has been tackled effectively by trained and seasoned human operators(drivers) but with the advent of self-driving cars, it is essential to automate real-time detection of these road surface deformities for optimal vehicle performance and safety of passengers. 

Pothole and Crack detection approaches can be further sub-classified into three categories, 3D Scanning\cite{i1}, vibration techniques\cite{i2} and Vision-based methods\cite{i3}\cite{i4}. The prohibitive cost of LIDARs limits the adaptability of the first category approaches whereas reliability in presence of vibrating surfaces like bridges has plagued the second category. In comparison, cameras are cheaper, ubiquitous and the current vision-based systems have enjoyed higher applicability and are more robust in wide-ranging scenarios, albeit suffering from a higher false positive rate. Traditionally, handcrafted filters\cite{i6}\cite{i7}have found widespread use in popular algorithms as they perform well on the acquired datasets, but they lack the discriminative powers to differentiate between the deformities and background noise at low-level resolutions. Other detection schemes use a combination of gradient features for each pixel followed by binary classification for prediction. Crack detection using Local binary patterns\cite{i8} and Gabor filters\cite{i9} have been proposed. In Crack-It\cite{i10},an unsupervised fully integrated system for crack detection and characterization has been proposed. Crack-It performs well for different crack types and also gives the average characteristics of the cracks detected.In \cite{i11} ,another unsupervised technique for pothole detection has been proposed based on image analysis and spectral clustering. they give a rough estimate of potholes along with the surface. In \cite{i12} a novel crack detection paradigm using convolution neural nets leveraging the discriminative powers of deeply learned features has been proposed.It does not make any assumptions regarding the geometry of the road unlike previous approaches which require optical axis of the camera be perpendicular to the road surface. Majority of these techniques perform poorly against complex backgrounds and the remaining approaches fail when the images are captured from a mobile platform making these unsuitable for application on intelligent transport systems.

\begin{figure}[t!]
\centering
\includegraphics[width=0.85\linewidth]{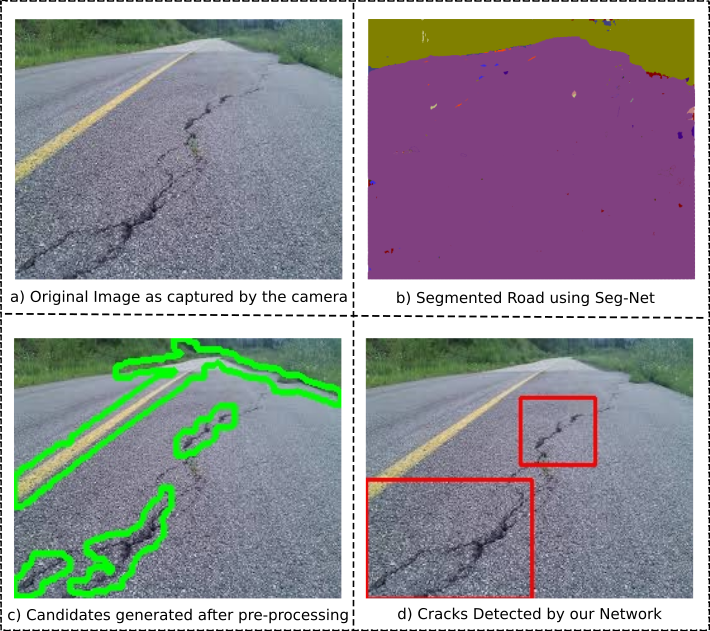}
\caption{Finding cracks and potholes over large viewpoint changes in the presence of noise, shadows, and occlusion is a tedious task. Our proposed method accurately detects road cracks and potholes using SegNet for road segmentation and other pre-processing procedures to generate candidates highly probable to be detected as cracks. Finally, our network a modified combination of Squeeze-Net and Encoding layer classify these candidates based on spatial and texture features.}
\end{figure}

\begin{figure*}[h!]
    \centering
\includegraphics[width=0.95\linewidth,height=0.2\linewidth]{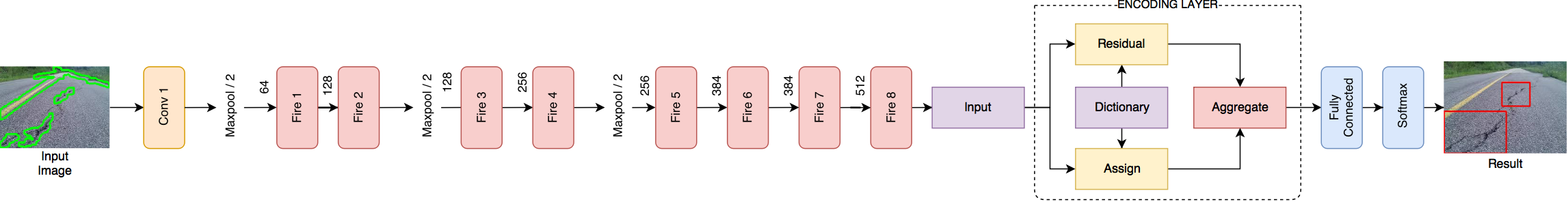}
    \caption{Architecture: SqueezeNet and Encoding Layer, We replaced the final convolution and global pooling layer with our encoding layer followed by a fully connected layer to act as a classifier. }
    \label{basic}
\end{figure*}

In this paper, we present a compact crack and pothole detection system designed for use on self-driving platforms to enable them to identify impending hazards. We propose a convolution neural net architecture for discriminating potholes and cracks from the background. Texture-based features learned in the encoding layer in conjugation with spatial information form the basis of classification. The major contributions of our paper can be listed as follows:-
\begin{itemize}
\setlength{\itemsep}{5pt}
\item A deep architecture which learns texture along with the spatial information. Texture features are important because we cannot fully rely on spatial information on classification as it is bound to fail when viewpoints change. Texture features remain consistent for different viewpoints and hence are better suited for the problem.
\item A pipeline which detects cracks and potholes in real time, ensured by our rigorous pre-processing methodology which provides probable candidates to be further analyzed by our network while most traditional approaches directly apply CNN to high-resolution images using sliding window, rendering these techniques unusable for real-time applications.
\item A robust classifier which is able to correctly and consistently differentiate between background noise,  shadows, cracks and various other ambiguous scenarios which cannot be differentiated using traditional approaches.
\item The proposed system can be used as a plug and play module in various autonomous robots and self-driving cars without changing their base design.
\end{itemize}
\vspace{3pt}
 The efficacy of the proposed method is evaluated on the following standard public datasets\cite{i12}\cite{i13}, and in result section, we compare our system against the state of the art techniques where it is shown that our system outperforms all the other competing techniques.
The rest of the paper is organized as follows:- proposed method, preprocessing and system architecture are described in section \label{pro}, followed by Experiments, results, detailed analysis and failure cases in \label{exp}. In section\label{con}, we conclude with final remarks.
\section{Proposed Method}\label{pro}
We propose an autonomous crack and pothole detection system. The system is based on deep neural network architecture which discriminates between road and potholes using texture-based features. The novelty of the approach lies in using texture as a basis to differentiate between road filled with cracks and road free from cracks. 
The architecture of the system is shown in Fig \ref{basic}. It consists of a convolution neural network which learns the spatial information from the images and a texture encoding layer to discriminate between images on the basis of their texture. We now describe our system and architecture.
\subsection{System}
Our system can be easily deployed on processing boards with an image capturing device attached to it, The board must support CUDA processing for real-time detection. In our setup, we mounted an IDS uEye LE camera on NVIDIA Jetson TX-2 board for real-time performance on a 640x480 resolution image at 30 Fps.
\begin{figure*}[h!]
    \centering
    \includegraphics[width=0.95\linewidth]{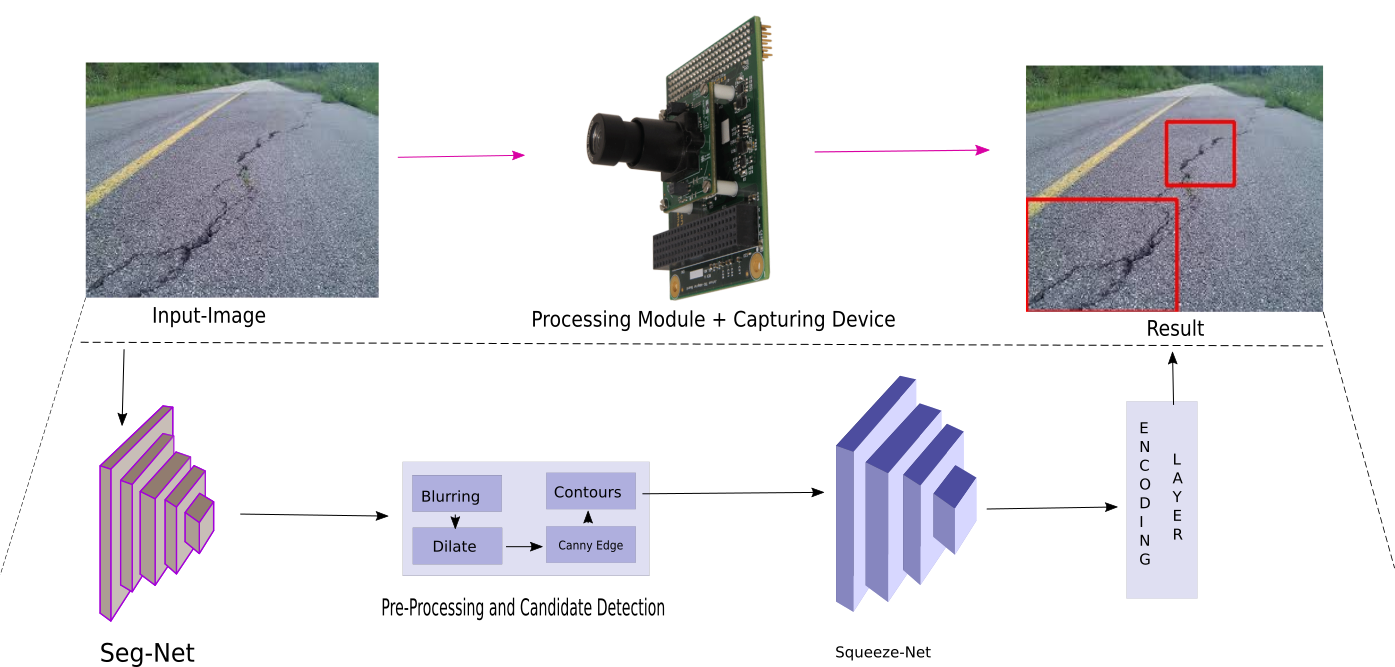}
    \caption{Full System Overview: The board mounted camera is used as a capturing device and the board is used as a processing module. }
    \label{system}
\end{figure*}
\subsection{Pre-processing}
The frames of a video are quite different from the training set and hence cannot be directly fed into the model for testing. So to bring the frame of the video to our domain, we follow the pre-processing steps as shown in Fig \ref{system}. Running the model directly on the segmented road would be inefficient leading to non-real time performance. To combat this we generate candidate regions from the road which are highly probable to be cracks and potholes. This whole procedure involves traditional image processing techniques to detect cracks. 
To generate candidate potholes and cracks we create two masks from the image, the first mask is created by passing the video frame through the modified SegNet\cite{i14}\cite{i15}\cite{i16}, which segments road from the scene, as we are not concerned with other parts in the image.
To create the second mask, a simple differentiation-based edge detection algorithm (Canny edge detection) is performed on the video frame. The Canny edge detector produces a 2-channel image with lighter pixels as edges and dark background. But the edges detected are unconnected. To connect these edges dilation is performed several times, which increases the area of the lighter pixels. 
Finally, the two generated masks are combined(AND) and contour detection is applied to the resultant mask. The bounding boxes of these contours are extracted from the input image which forms our candidates for suspected potholes and cracks. We resize these candidate regions into 64x64 patches, which are then passed through our deep neural net architecture.

Unwanted edges, especially around the outer boundaries of the road, are created by shadows of branches and leaves in trees and are more often worsened by sections of light shining through. Additionally, other vehicles on the road also create unwanted edges. These unwanted edges end up existing as false candidates for our networks. 

\subsection{Architecture}
We derive our architecture Fig \ref{basic}. from widely popular SqueezeNet\cite{i17}.SqueezeNet has a 1.4x smaller model size than AlexNet\cite{i18} while maintaining or exceeding the baseline accuracy of AlexNet.For a typical 64x64 image patch it provides 30x speedup as compared to AlexNet, which helps us to achieve real-time performance. SqueezeNet achieves the above claims by introducing a $Fire module$ which consists of a squeeze convolution layer(which has only 1x1 filters), feeding into an expand layer that has a mix of 1x1 and 3x3 convolution filters. The 1x1 filters in Fire modules help to reduce size as of the model since a 1x1 filter has 9X fewer parameters than a 3x3 filter. We remove the last convolution layer of the SqueezeNet and incorporate an encoding layer which is described in the next section. The output of the encoding layer is fed into a fully connected layer which is used for classification. Introduction of the encoding layer helps to learn the texture features alongside the spatial features.

\subsection{Encoding layer}
The encoding layer adds dictionary learning and residual encoding to the network through a single layer of CNN.\cite{i19}.It creates a dictionary by assigning each descriptor to $K$ codewords using some weights.The Encoding Layer acts as a pooling layer by encoding robust residual representations, which converts arbitrary input size to a fix length representation.The encoding layer maps each of the N feature vectors $F = \big\{f_1, ..f_N \big\}$ to K codewords $C = \big\{c_1, ...c_K\big\}$ with $a_{iK}$ as the assignment weights. Given $r_{ik} = f_i - c_k$ and $s_k$ as smoothing factor for each cluster center $c_k$ the assignment weights are given by 
\begin{equation}
a_{ik} = \frac{\exp{(-s_k \Vert {ik} \Vert^2)}}{\Sigma_{j=1}^k\exp{(-s_j \Vert r_{ik} \Vert ^2)}}
\end{equation}
\subsection{Training}
Given a training set $S = \big\{ x^{(i)} , y^{(i)} \big\} $ which contains m image patches, where $x^{(i)}$ is the i-th image patch and $y^{(i)} \in \big\{ 0, 1 \big\}$ is the corresponding class label. If $y^{(i)} = 1$, then $x^{(i)}$ is a positive patch, otherwise $x^{(i)}$ is a negative patch. All convolution filter kernel elements are trained from the data in a supervised fashion by learning from the labeled set of examples. The input images is passed through the network, which produces features,Finally a fully connected layer is used for classification. Softmax layer is used as last layer of the network as the classes are mutually exclusive. We use binary cross entropy loss as our loss function for training.
\footnote{https://github.com/sukhad-app/Crack-Pot.git}
\begin{figure*}[h!]
    \centering
    \includegraphics[width=0.95\linewidth,height=0.55\linewidth]{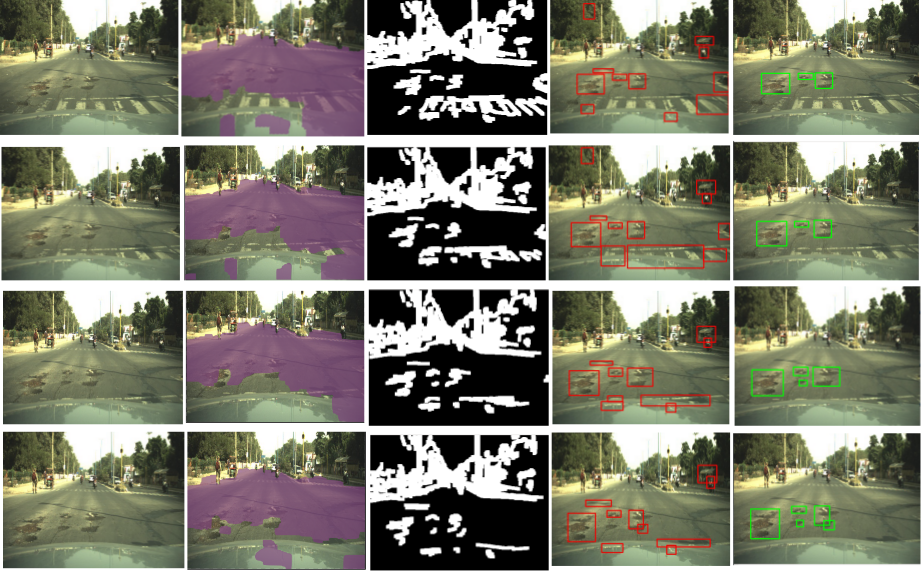}
    \caption{Processing on consecutive frames of a video }
    \label{result}
\end{figure*}

\section{Experiments}\label{exp}
All the experiments have been conducted on a workstation with 1.2 GHz CPU, 32 GB RAM, NVIDIA P5000 GPU, running on Ubuntu 14.04. In these experiments, we use Adam optimizer with a batch size of $64$, momentum of 0.9 and a learning rate of $0.00001$ and trained the models for 20 epochs. For the experiments, we keep the value of $K$ as 32. To test the speed and accuracy of our model in real time scenarios, we produce results on the videos captured through our system mounted on a moving car. We train our model on two different datasets and compare our accuracy with the baseline accuracy reported on them by earlier approaches.
\subsection{Datasets}
We report our accuracy on two available datasets. All other publicly available datasets have less than 300 images combined\cite{i7}\cite{i10}, hence they cannot be used for training our model.
\paragraph*{GAPs dataset}
The GAPs dataset includes a total of 1,969 gray valued images (8bit), partitioned into 1,418 training images, 51 validation images, and 500 test images. The image resolution is 1920x1080 pixels with a per pixel resolution of 1.2mm x 1.2mm. The pictured surface material contains pavement of three different German federal roads. These images are divided into 64x64 patches and each patch is labeled as a crack or not. We train our model on 50 training chunks from the 154 available training chunks. Each chunk contains 32000 64x64 images where each image either contains a crack or is crack-free. 

\paragraph*{Zhang dataset}
This dataset contains 700 pavement pictures of size 3264× 2448 collected at the Temple University campus using a smartphone as the data sensor. Each image is annotated by multiple annotators. These images are further divided into patches where each patch either contains a crack or is crack-free. Each sample is a 3-channel (RGB) 99×99 pixel image patch generated by the sampling strategy described by Zhang et al. in \cite{i12} There are total 2 million available annotated samples, where each sample either contains a crack or is crack free. We train our model on around 1.3 million samples and use the rest for testing and validation.

\subsection{Results}
Precision and Recall are defined as:
\begin{equation}
P = \frac{True Positive}{False Positive + True Positive} 
\end{equation}
\begin{equation}
R = \frac{True Positive}{True Positive + False Negative}
\end{equation}
and F1 score as :
\begin{equation}
F1 = \frac{2PR}{P + R}
\end{equation}
To compare our results with GAPs dataset we use accuracy as the measure which is defined as:
\begin{equation}
Accuracy = \frac{TP + TN}{TP + TN + FP + FN}
\end{equation}
We use these metrics to compare our results with the baselines. Table \ref{ICIP} compares our results on the dataset with the baselines given by Zhang et al. in \cite{i12}. It shows that our approach performs much better than traditional approaches and the deep architecture given by the authors. Table \ref{GAPs} depicts that our approach even outperforms recent deep neural networks that have been developed. 
\begin{table}[h]
\caption{Comparison with ICIP dataset}
\label{ICIP}
\begin{center}
\begin{tabular}{|c||c||c||c|}
\hline
Approach&Precision & Recall & F1 score\\
\hline
SVM & 0.8112&0.6734&0.7359\\
Boosting&0.7360&0.7587&0.7472\\
ICIP-Conv-net&0.8696&0.9251&0.8965\\
\textbf{Ours}&\textbf{0.9237}&\textbf{0.9376}&\textbf{0.9301}\\
\hline
\end{tabular}
\end{center}
\end{table}
\begin{table}[h]
\caption{Comparison with GAPs dataset}
\label{GAPs}
\begin{center}
\begin{tabular}{|c||c||c||c|}
\hline
Approach&Accuracy&F1 score\\
\hline
\textbf{Ours}& \textbf{0.9893}&0\textbf{0.7314}\\
ASIVNOS net&0.9772&0.7246\\
ASIVNOS-mod&0.9723&0.6707\\
RCD net&0.9732&0.6642\\
\hline
\end{tabular}
\end{center}
\end{table}

\subsection{Results on generic scenes}
As we have developed a system which can work on scenes which the robot or an autonomous vehicle may encounter while driving. So we also present our results on generic scenes that a robot may encounter. Fig. \ref{result} represents the results of our proposed testing pipeline on consecutive frames of a video. The image shows processing of 5 consecutive frames captured through a camera mounted on the dashboard of the car. The images are outputs of the following respectively from left to right: Input image, Input image after processing using SegNet\cite{i14}\cite{i15}\cite{i16}, Input image after edge detection and dilation, The candidate potholes generated after combining the SegNet and preprocessing masks, Final candidates detected by our network as prospective potholes and cracks. Note that the results are consistent across various frames. Also, it is quite evident from the images that our architecture is easily able to remove all the false positives that it may encounter after the preprocessing. The shadows generated which may seem as a pothole have been removed.  

\subsection{Detailed Analysis}
As we see above our architecture performs better than all traditional approaches and deep neural architectures developed in the past. This depicts that taking texture features into account, along with the spatial features make the classifier robust. If we analyze our results on the generic dataset, we see that our system gives consistent results on consecutive frames of a video and is easily able to remove all potential false positives like shadows etc. This shows that our system is robust towards change in viewpoint. 
\subsection{Failure Cases}
\begin{figure}[h!]
\centering
\includegraphics[width=0.45\linewidth,height=0.35\linewidth]{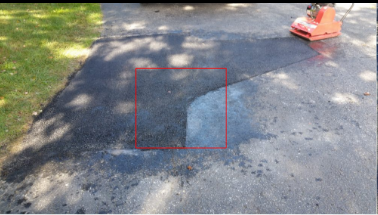}
\includegraphics[width=0.45\linewidth,height=0.35\linewidth]{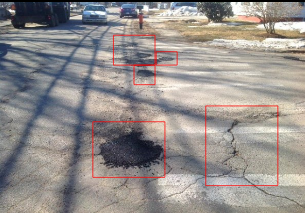}
    \caption{Failure Cases}
    \label{faliure}
\end{figure}
The above images exemplify the typical failure cases which result from the structural ambiguity between surface deformities and restoration patches. Generally, the texture of the restoration patches and repairs are too dissimilar to the texture of the actual road such that classifying them as cracks or potholes becomes challenging even for humans. Such cases end up giving false positives.

\section{CONCLUSIONS}\label{con}

We proposed a novel system for autonomous crack and pothole detection which performs in real-time and is able to handle large viewpoint changes, background-noise, partial occlusion, and shadows, giving robust and consistent results on datasets and real-life videos as well. A significant increase in accuracy is observed on standard road crack datasets which proves the robustness and efficacy of the approach. The approach performs exceptionally well on live videos but fails at restoration patches, which are quite difficult to ascertain even with human brain due to lack of knowledge about depth. Further, various navigational strategies like A*, PRM can be applied to evade the detected potholes. The approach can also be used by Municipal authorities to examine road conditions with the help of autonomous bots. 

\addtolength{\textheight}{-12cm}   


\begin{thebibliography}{99}
\bibitem{i1} K.T. Chang, J.R. Chang, J.K. Liu, Detection of pavement distresses using 3D laser scanning technology, Computing in Civil Engineering, pp. 1-11, 2005
\bibitem{i2} J. Eriksson, L. Girod, B. Hull, R. Newton, S. Madden, H. Balakrishnan, The pothole patrol: using a mobile sensor network
for road surface monitoring, Proceeding of the 6th international conference on Mobile systems, applications, and services, pp.
29-39, 2008
\bibitem{i3} G. M. Jog, C. Koch, M. Golparvar-Fard, I. Brilakis, Pothole Properties Measurement through Visual 2D Recognition and 3D
Reconstruction, Computing in Civil Engineering, pp. 553-560, 2012
\bibitem{i4} C. Koch, I. Brilakis, Pothole detection in asphalt pavement image, Advanced Engineering Informatics, Vol. 25(3), pp. 507-
515, 2011 
\bibitem{i6} S. Varadharajan, S. Jose, K. Sharma, L. Wander, and C. Mertz, “Vision for road inspection,” in Proceedings of 2014 IEEE Winter Conference on Applications of Computer Vision, 2014, pp. 115–122.
\bibitem{i7} Q. Zou, Y. Cao, Q. Li, Q. Mao, and S. Wang, “Cracktree: Automatic crack detection from pavement images,” Pattern Recognition Letters, vol. 33, no. 3, pp.227–238, 2012.
\bibitem{i8} Y. Hu and C. Zhao, “A local binary pattern based methods for pavement crack detection,” Journal of Pattern
Recognition Research, vol. 5, no. 1, pp. 140–147, 2010.
\bibitem{i9} M. Salman, S. Mathavan, K. Kamal, and M. Rahman,“Pavement crack detection using the gabor filter,” in
Proceedings of IEEE International Conference on Intelligent Transportation Systems, Oct. 2013, pp. 2039–2044.
\bibitem{i10} H. Oliveira and P. L. Correia, "Automatic Road Crack Detection and Characterization," in IEEE Transactions on Intelligent Transportation Systems, vol. 14, no. 1, pp. 155-168, March 2013.
doi: 10.1109/TITS.2012.2208630
\bibitem{i11} Buza, E.; Omanovic, S.; Huseinnovic, A. Pothole detection with image processing and spectral clustering. In Proceedings of the 2nd International Conference on Information Technology and Computer Networks, Antalya, Turkey, 8–10 October 2013; pp. 48–53.
\bibitem{i12} L. Zhang, F. Yang, Y. Daniel Zhang and Y. J. Zhu, "Road crack detection using deep convolutional neural network," 2016 IEEE International Conference on Image Processing (ICIP), Phoenix, AZ, 2016, pp. 3708-3712.
doi: 10.1109/ICIP.2016.7533052
\bibitem{i13}Eisenbach, Markus and Stricker, Ronny and Seichter, Daniel and Amende, Karl and Debes, Klaus and Sesselmann, Maximilian and Ebersbach, Dirk and Stoeckert, Ulrike and Gross, Horst-Michael,"How to Get Pavement Distress Detection Ready for Deep Learning? A Systematic Approach," 2017 International Joint Conference on Neural Networks (IJCNN) pp. 2039--2047.
\bibitem{i14}Kendall, Alex and Badrinarayanan, Vijay and and Cipolla, Roberto, Bayesian SegNet: Model Uncertainty in Deep Convolutional Encoder-Decoder Architectures for Scene Understanding, arXiv preprint arXiv:1511.02680, 2015
\bibitem{i15}Badrinarayanan, Vijay and Kendall, Alex, and Cipolla, Roberto, SegNet: A Deep Convolutional Encoder-Decoder Architecture for Image Segmentation, IEEE Transactions on Pattern Analysis and Machine Intelligence, 2017
\bibitem{i16}Badrinarayanan, Vijay and Handa, Ankur, and Cipolla, Roberto, SegNet: A Deep Convolutional Encoder-Decoder Architecture for Robust Semantic Pixel-Wise Labelling, arXiv preprint arXiv:1505.07293, 2015
\bibitem{i17}Forrest N. Iandola and Song Han and Matthew W. Moskewicz and Khalid Ashraf and William J. Dally and Kurt Keutzer, SqueezeNet: AlexNet-level accuracy with 50x fewer parameters and $<$0.5MB model size, arXiv:1602.07360, 2016
\bibitem{i18}Alex Krizhevsky and Ilya Sutskever and Geoffrey E. Hinton, Imagenet classification with deep convolutional neural networks, Advances in Neural Information Processing Systems, 2012, pp. 1097--1105
\bibitem{i19}Zhang, Hang and Xue, Jia and Dana, Kristin,Deep TEN: Texture Encoding Network,arXiv preprint arXiv:1612.02844,
2016
\end{thebibliography}
\end{document}